%% file: main.tex
\documentclass[conference]{IEEEtran}
\IEEEoverridecommandlockouts
\usepackage[utf8]{inputenc} % allow utf-8 input
\usepackage[T1]{fontenc}    % use 8-bit T1 fonts
\usepackage{hyperref}       % hyperlinks
\usepackage{url}            % simple URL typesetting
\usepackage{booktabs}       % professional-quality tables
\usepackage{amsfonts}       % blackboard math symbols
\usepackage{nicefrac}       % compact symbols for 1/2, etc.
\usepackage{microtype}      % microtypography
\usepackage{graphicx}			
\usepackage{epsfig}
\usepackage{floatrow}
\usepackage{tabu}
\usepackage{cite}
\usepackage{amsmath,amssymb,amsfonts}
\usepackage{algorithmic}
\usepackage{textcomp}
\usepackage{xcolor}
\usepackage{amsmath}
\usepackage[ruled]{algorithm2e}
\algsetup{linenosize=\small}
\def\BibTeX{{\rm B\kern-.05em{\sc i\kern-.025em b}\kern-.08em
    T\kern-.1667em\lower.7ex\hbox{E}\kern-.125emX}}

\usepackage{graphics} % for pdf, bitmapped graphics files
\usepackage{times} % assumes new font selection scheme installed
\usepackage{color}
\usepackage{multirow}

\usepackage{xspace}
\makeatletter
\DeclareRobustCommand\onedot{\futurelet\@let@token\@onedot}
\def\@onedot{\ifx\@let@token.\else.\null\fi\xspace}
\let\labelindent\relax

\newcommand{\Cref}[1]{Chap.~\ref{#1}}

\author{%
 Xin Shen, Praful Agrawal, Zhongwei Cheng \\}

\begin{document}
\title{Data Efficient Training with Imbalanced Label Sample Distribution for Fashion Detection}
\maketitle

%=================================
%abstract
%=================================
\input{sec0_abstract}

%================================
%introduction
%================================
\input{sec1_introduction}
\input{sec2_relatedwork}

%==============================
%Technical Approach
%=============================
\input{sec3_technicalapproach}

%============================
%Experiment Results
%============================
\input{sec4_exp}

%===========================
%Conclusion
%===========================
\input{sec5_conclusion}

\bibliographystyle{plain}
\bibliography{references}

% \section*{Appendices}
\appendix
\section{Mathematical Proof of Effective Number of Samples}
\label{sec:appendix}
The Effective Number of Samples is proved by induction. Suppose we have sampled n - 1 examples for a class currently, and we have not yet sample the $n^{th}$ sample. Now, suppose, the expected volume of the previously sample data is $E_{n-1}$ , then the newly sampled data point holds a probability of $p = \frac{E_{n- 1}}{N}$ in order to be overlapped with the previously sampled data samples. Then, the expected volume after sampling the $n^{th}$ example is :
\begin{equation}
    E_n = p*E_{n - 1} + (1 - p)*(E_{n - 1} + 1) = 1 + \frac{N - 1}{N}E_{n - 1}
\end{equation}
If we assume $E_{n - 1} = \frac{1 - \beta^{n - 1}}{1 - \beta}$ is valid, then:
\begin{equation}
    E_n = 1 + \beta\frac{1 - \beta^{n - 1}}{1 - \beta} = \frac{1 - \beta + \beta - \beta^{n}}{1 - \beta} = \frac{1 - \beta^n}{1 - \beta}
\end{equation}

% \section{Label Distribution Plot Illustration of Archetype}

\end{document}

%% file: sec0_abstract.tex
\begin{abstract}
% Multi-label classification models have broad applications in E-commerce production features, from visual based label predictions to language based sentiment classifications. One challenge to achieve satisfactory performance for those classification tasks in real world is the notable imbalanced data distribution; for example, for fashion attribute detection, you can find only 6 `puff sleeve' clothes by browsing 1000 products in most E-commerce's Fashion catalog. To enable our model accurately predicting those under-represented labels, we either acquire huge amount of annotations to collect sufficient samples, which is neither economic nor scalable, or we explore more data efficient model training techniques. In this paper, we employ a state-of-the-art weighting of objective function to boost deep neural network's (DNN) performance for multi-label classification with long-tailed data distribution. Experiments involve image based attribute classification of fashion apparels. Results show favourable performance for the new weighting method in comparison to non-weighted and inverse-frequency based weighting mechanism. We further evaluate the robustness of the new weighting mechanism using two fashion attributes -- sleevetype and archetype.

Multi-label classification models have a wide range of applications in E-commerce, including visual-based label predictions and language-based sentiment classifications. A major challenge in achieving satisfactory performance for these tasks in the real world is the notable imbalance in data distribution. For instance, in fashion attribute detection, there may be only six 'puff sleeve' clothes among 1000 products in most E-commerce fashion catalogs. To address this issue, we explore more data-efficient model training techniques rather than acquiring a huge amount of annotations to collect sufficient samples, which is neither economic nor scalable. In this paper, we propose a state-of-the-art weighted objective function to boost the performance of deep neural networks (DNNs) for multi-label classification with long-tailed data distribution. Our experiments involve image-based attribute classification of fashion apparels, and the results demonstrate favorable performance for the new weighting method compared to non-weighted and inverse-frequency-based weighting mechanisms. We further evaluate the robustness of the new weighting mechanism using two popular fashion attribute types in today's fashion industry: sleevetype and archetype.

\end{abstract}

%% file: sec1_introduction.tex
\section{Introduction}
%Outline: Amazon wide applications for multi-label classification, typical problem, some solutions, effective numbers, summary of work
%Within the Amazon of exciting machine learning problems, many classification tasks involve multiple labels. 
Data imbalance refers to the problem of uneven representation of class labels. Both computer vision and natural language processing applications involve the multi-label problems such as classification of natural images (CIFAR \cite{krizhevsky2012imagenet}), attribute tagging of fashion images \cite{gutierrez2018deep}, sentiment analysis \cite{kim2019sentiment}, and hierarchical multi-label text classification \cite{gargiulo2019deep}. With the advancement of convolutional neural network (CNNs)~\cite{goodfellow2016deep, goodfellow2016convolutional} and the era-break architecture of ResNet~\cite{he2016deep}, many related classification-applications in today's industry~\cite{shen2023semantic,deng2009imagenet,zhang2022resnest, shen2023fishrecgan} have shown a significant robustness. However, none of these model developments could really resolve the imbalanced data issue. Common solutions include alternative models such as one-class learning for a binary classification, data augmentation by way of linear transformations on existing data~\cite{liu2021two}, prior-based modeling such as weighted loss mechanism to synthetically boost the representation of minority classes. Data augmentation helps fill in the gaps of underlying training data distribution, however, it is still limited to address the skewed sample sizes. Most methods chose to weight the label-specific loss with a smoothed version of inverse-frequency of that label \cite{elkan2001foundations, huang2016learning, lin2014microsoft, malisiewicz2011ensemble, wah2011caltech}. Though effective in some cases, the improved performance of minority class often occurs at the cost of majority class performance. In this work we investigate a newly proposed weighting scheme \cite{cui2019class} for tagging the fashion-based attributes in fashion apparel images. The experimental results showcase the success and limiting scenarios of the method. Next section briefly presents the details of weighting scheme by Cui et al. \cite{cui2019class}, followed by a discussion of experimental design and results.

%% file: sec2_relatedwork.tex
\section{Related Work}
%@Praful, please refer this part when you want to merge the information into Intro Sec
\subsection{Naive Weighted Loss}
A general approach to assign the weightings/ costs $\in \mathbb{R}^n$ of n classes for each class is the inverse proportion weightings: set the inverse of each class's support or as the inverse of the square root of each class's support as the weightings. Then, usually, we need to ensure that for minority classes, the weightings/ costs should be above 1. Thus, usually, we would multiply the inverse proportion weightings with a $\mathbb{\lambda}$  multiplier as the scaling parameter to ensure the weighting before minority classes are above 1. 
\begin{equation}
    weights_{i} = \frac{1}{Label Support_{i}} \cdot \lambda, i \in \mathbb{R}^n
\end{equation}

Once the inverse proportion weightings are calculated, the weighted loss can be averaged across observations.
\begin{equation}
    \mathcal{WL}(x, class) = weights_{class} \cdot \mathcal{L}[class]
\end{equation}

\begin{equation}
    \mathbb{L} = \frac{\sum_{i = 1}^{n} \mathcal{WL}(class[x, i])}{\sum_{i = 1}^{n} weights(class[i])}
\end{equation}
However, we can see from the equations above, the weighting vector calculated can only represent the current sample population, which fails to take the whole population's information into consideration; that is, based on the current dataset, we fail to infer what the sample population would be distributed if more samples are added or to infer what the true population would be.

%% file: sec3_technicalApproach.tex
\section{Technical Approach}

In order to address the imbalance in label distribution, the idea of weighted loss has been explored in various applications with limited success. Commonly the label weights are derived from inverse of label frequency. This approach has a fundamental assumption that the given number of samples are needed to represent the class distribution. One of the ways to resolve this limitation is to acquire additional samples. However, due to the inherent information overlap among data, the improvement in model performance is marginal. Cui et al. \cite{cui2019class} address this limitation by estimating the "effective number" of samples to represent a given data distribution. The inverse of this effective sample size is then used to estimate a class-balanced loss function. The objective is to reduce information overlap among data, while computing the effective sample size. 

\begin{figure}[!h]
    \centering
    \includegraphics[width=0.7\textwidth]{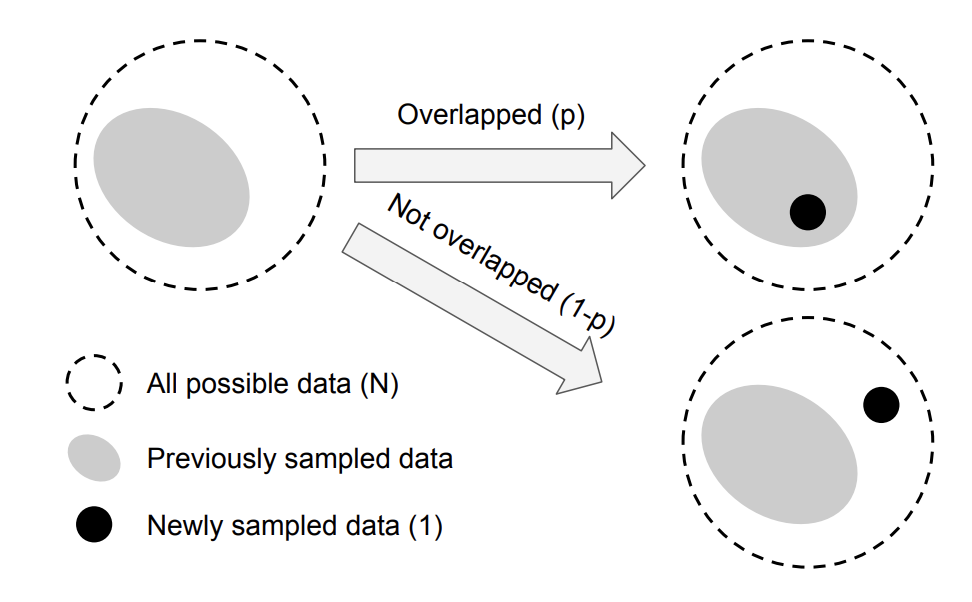}
    \caption{The illustration from Cui et al. \cite{cui2019class}. Given the full set of all possible data with volume $N$ and a label with size $n$, a newly introduced sample for that label holds a probability $p$ to be overlapped with the existing distribution, and $(1-p)$ not to be overlapped.}
    \label{fig:overlapped}
\end{figure}

The idea of effective numbers is based on random sampling the data to cover the sample distribution. The effective number of samples ($E_n$) can be imagined as the actual volume or the expected actual volume of samples for a class which suggests that %proposed a thinking that %based on the current sampled data, 
if we continue to augment more data points, many newly collected data points could either be ``overlapped" or ``not overlapped" with the current sample distribution \cite{cui2019class} . The Bernoulli probability $p$ represents the chance of ``overlapped" and thus $(1-p)$ represents the ``not overlapped", as shown in Figure~\ref{fig:overlapped}. Using this probabilistic model, the effective number of samples for the $i$--th label with total $n_i$ samples can be computed as $E_{n_{i}} = \frac{1 - \beta^{n_{i}}}{1- \beta}, i \in \{1,\cdots,K \}$. The hyperparameter $\beta \in [0,1)$ is derived from the possible sample space of each label with size $N$ as, $\beta =\frac{(N - 1)}{N}$. The effective number $E_n$ equals $n$ as the beta value approaches 1. We use $\beta = 0.99$ for our experiments. In our multi-label classification experiments, the weighted $CrossEntropy$ loss functions are used. The effective weights $(\frac{1}{E_{n_i}})$ are normalized to sum to 1. The weighted loss is obtained through the class-wise multiplication of the effective weights.
% \begin{equation}
%     \label{eq:effective}
%     E_{n_{i}} = \frac{1 - \beta^{n_{i}}}{1- \beta}, i \in \{1,\cdots,K \}
% \end{equation}

% \begin{equation}
%     \label{eq: beta}
%     \beta =\frac{(N - 1)}{N}
% \end{equation}
%Refer Appendix.\ref{sec:appendix} for a thorough mathematical proof of how the expression of the Effective Number of Samples are generated. Formally, by using the Bernoulli Distribution, authors of this previous work generated the expression of the Effective Number of Samples. With a class space $\|\mathbb{K}\|$, each label's volume is denoted as n shown in Eq. \ref{eq:effective}
%From the Eq. \ref{eq: beta}, it is clear that this hyper- parameter $\beta$ can be inferred as 0.9 or 0.999. Finally, we can treat the Effective Number of Samples as the expected label distribution. Similar to the previous introduced inverse proportion weightings, we can find the Effective Number of Samples based weighted loss expressed in Eq. \ref{eq: effloss}
%\begin{equation}
%    \label{eq: effloss}
%    \mathcal{WL}(x, class) = \frac{1}{W_{n_{class}}} * \mathcal{L}[class] 
%\end{equation}
%Up to now, we would take the inverse of the Effective Numbers of Sample, and the weighted loss can be averaged across observations.
% \begin{equation}
%     \mathbb{L} = \frac{\sum_{i = 1}^{n} \mathcal{WL}(class[x, i])}{\sum_{i = 1}^{n} weights(class[i])} \cdot \lambda
% \end{equation}

%% file: sec4_exp.tex
\section{Experiments}
We employ the effective number based weighted loss in multi-label image-based classification of two fashion attributes - archetype and sleevetype. In Fig.\ref{sec:appendix2}, it is clear that for the category of archetype, our data shows a significant imbalanced distribution. We obtained large amount of Fashion apparel images for women's shirts and sweaters are tagged with one sleevetype from \textit{\{balloon, batwing, bell, cap, dolman, fitted, puff, raglan, short, sleeveless, spaghetti\}}. The archetype attribute covers women's fashion images containing jumpsuit/rompers, dress, jeans, outerwear, pants, shoes, shorts, shirts, skirts, and sweater. Each image is tagged with one or more archetypes from \textit{\{androgynous, boho, casual, classic, edgy, glam, minimal, retro, romantic, sporty\}}. Next, we discuss the design details of the two models, followed by analysis of the observed results. We use the archetype model to compare non-weighted loss (referred as NW) with the two weighting mechanisms - inverse label frequency (referred as IFW), and inverse of effective size (IEW).

\label{sec:appendix2}
\begin{figure}[h]
    \centering
    \includegraphics[width=0.7\textwidth]{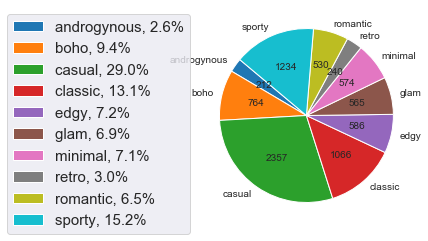}
    \caption{The data distribution visualization for Archetype dataset: After pre- filtering, we presented a correspondingly proper long- tail distribution with all minority classes accounting for 2.6\% to 7.1\%}
    \label{fig:archDist}
\end{figure}

\subsection{Design Details}
We apply Densenet \cite{huang2017densely} as backbone network for our multi-label classifiers of the two fashion attributes. The sleevetype model is trained against the non-weighted and effective-numbers-weighted $Softmax CrossEntropy$ loss functions using adam optimizer \cite{kingma2014adam}. The learning rate of $1e^{-5}$ resulted in stable training performance. We use top@1 mechanism to select the predicted label for sleevetype. 
The archetype model is trained against the weighted and non-weighted variants of the $Sigmoid CrossEntroy$ loss function using adam optimizer \cite{kingma2014adam}. The learning rate of $1e^{-4}$ resulted in stable training performance. We use independent thresholding mechanism to select the predicted labels for archetype. 

For both attributes, the complete datasets are randomly partitioned into training (90\%) and testing (10\%) sets. The model performance is evaluated using the area under the curve $(AUC)$ of precision-recall curve on test set.

\subsection{Results and Discussion}
\subsubsection{Major Result Analysis}
\begin{figure}[!h]
    \centering
    \includegraphics[width=0.47\columnwidth]{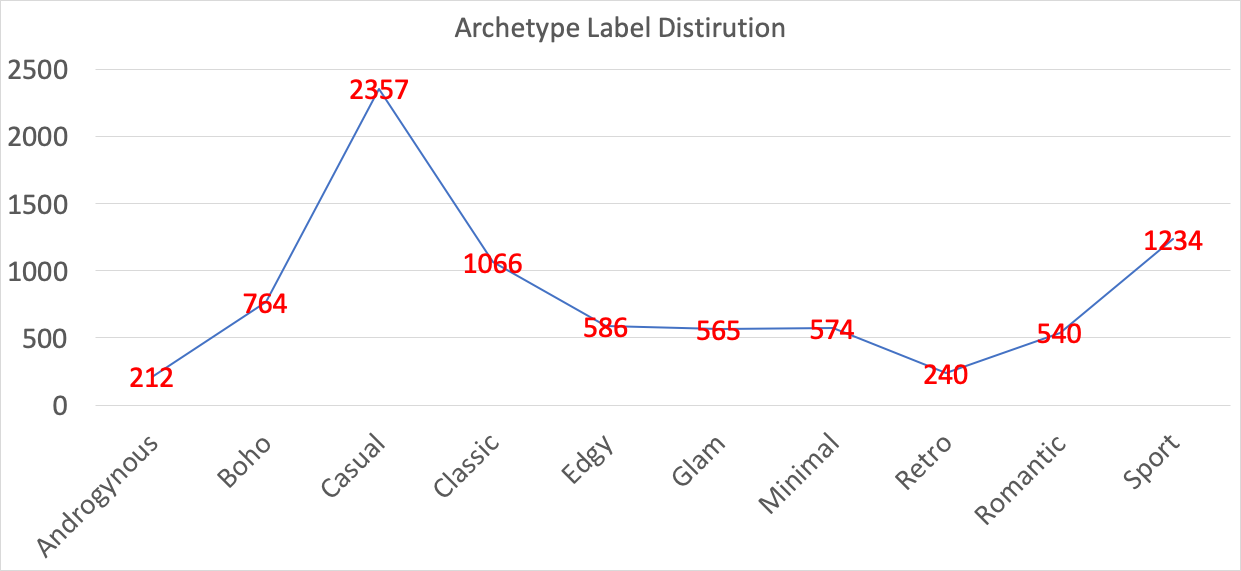}~~\includegraphics[width=0.5\columnwidth]{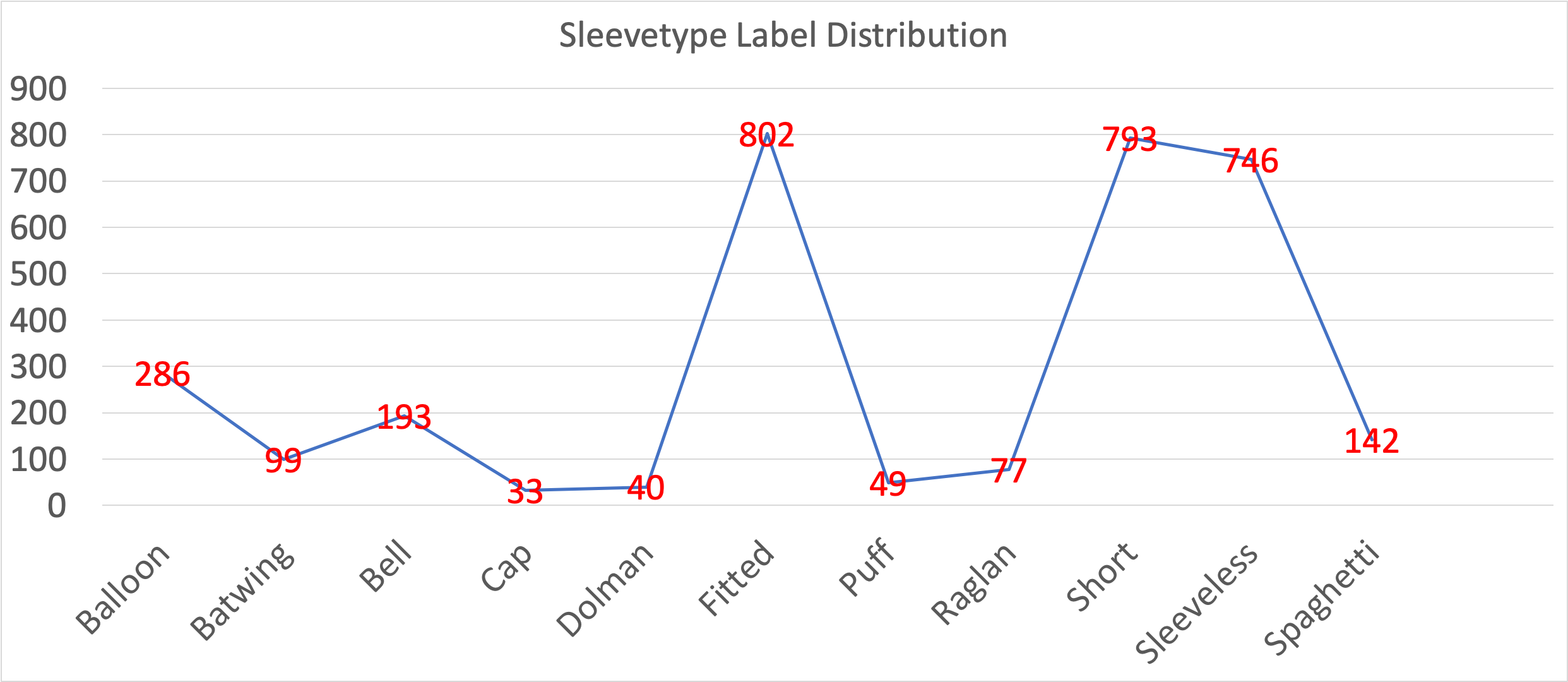}
    \caption{Data distributions of sleevetype and archetype datasets.}
    \label{fig:labelDist}
\end{figure}

Figure~\ref{fig:labelDist} shows the imbalanced label distributions in the two datasets. The archetype experiment showcases the advances of model trained with IEW. While the sleevetype dataset poses an additional challenge of extreme distribution with very low sample size (less than 100) for many labels, we present results for stress-testing the IEW model.

\begin{figure}[!h]
    \centering
    \includegraphics[height=1in]{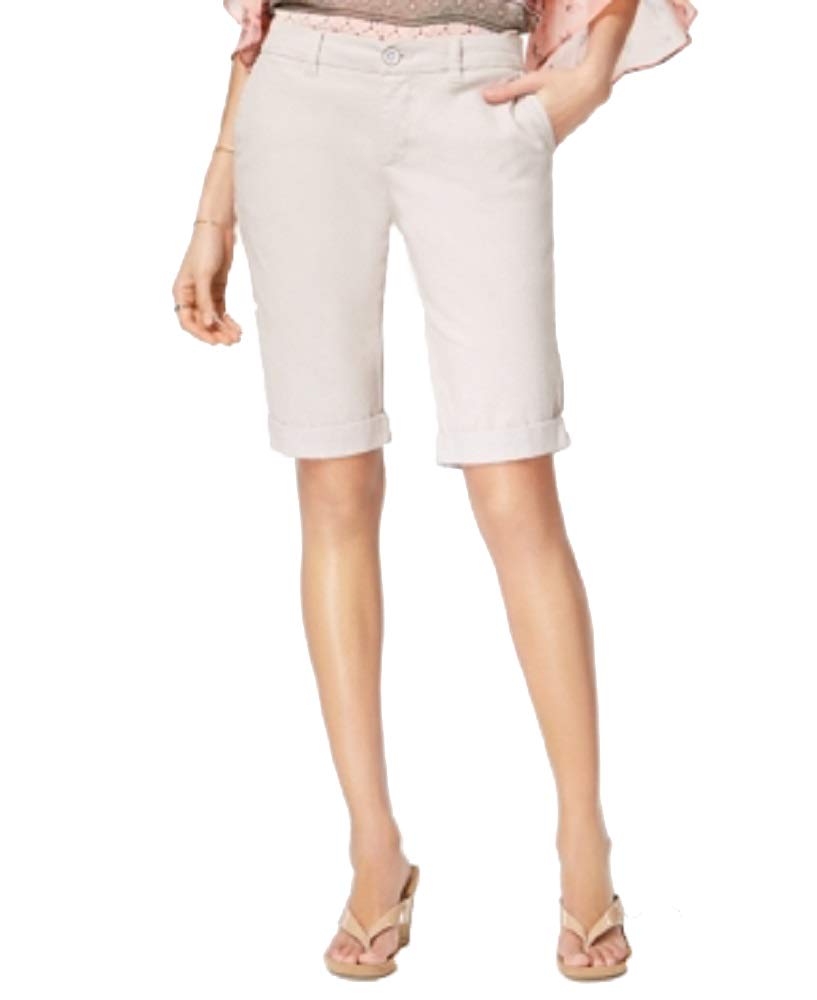}~~\includegraphics[height=1in]{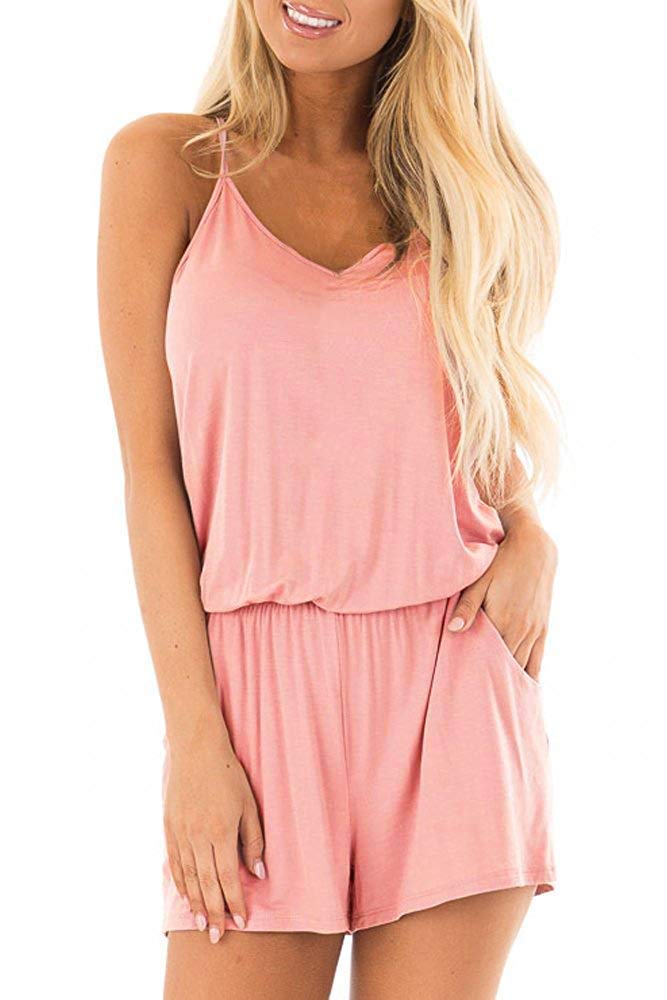}~~\includegraphics[height=1in]{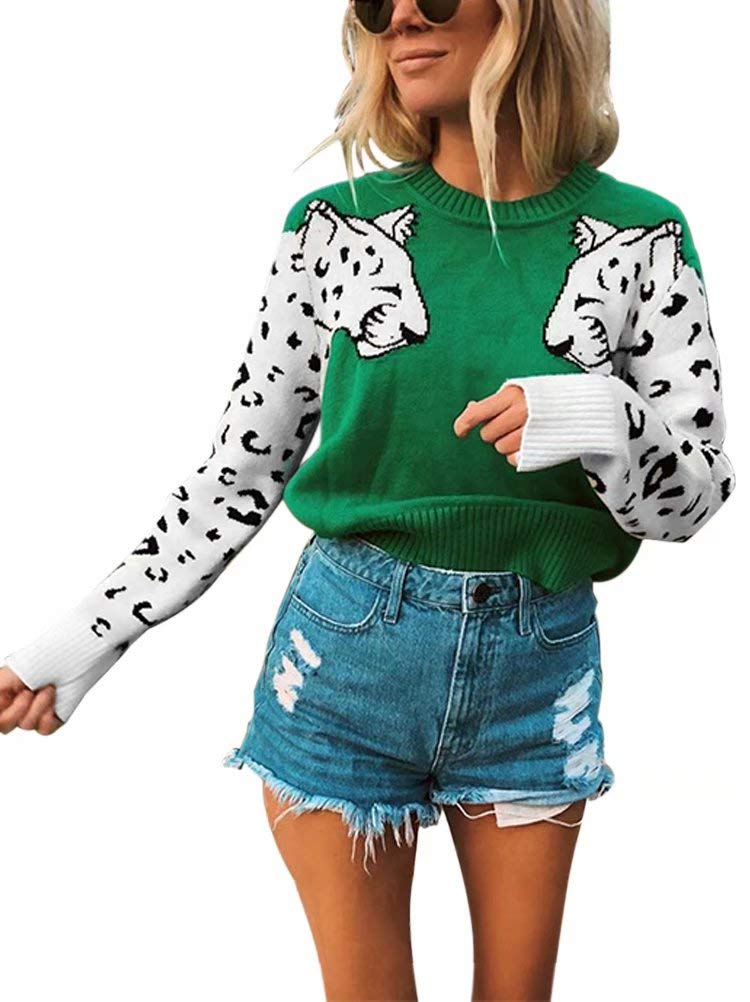}~~\includegraphics[height=1in]{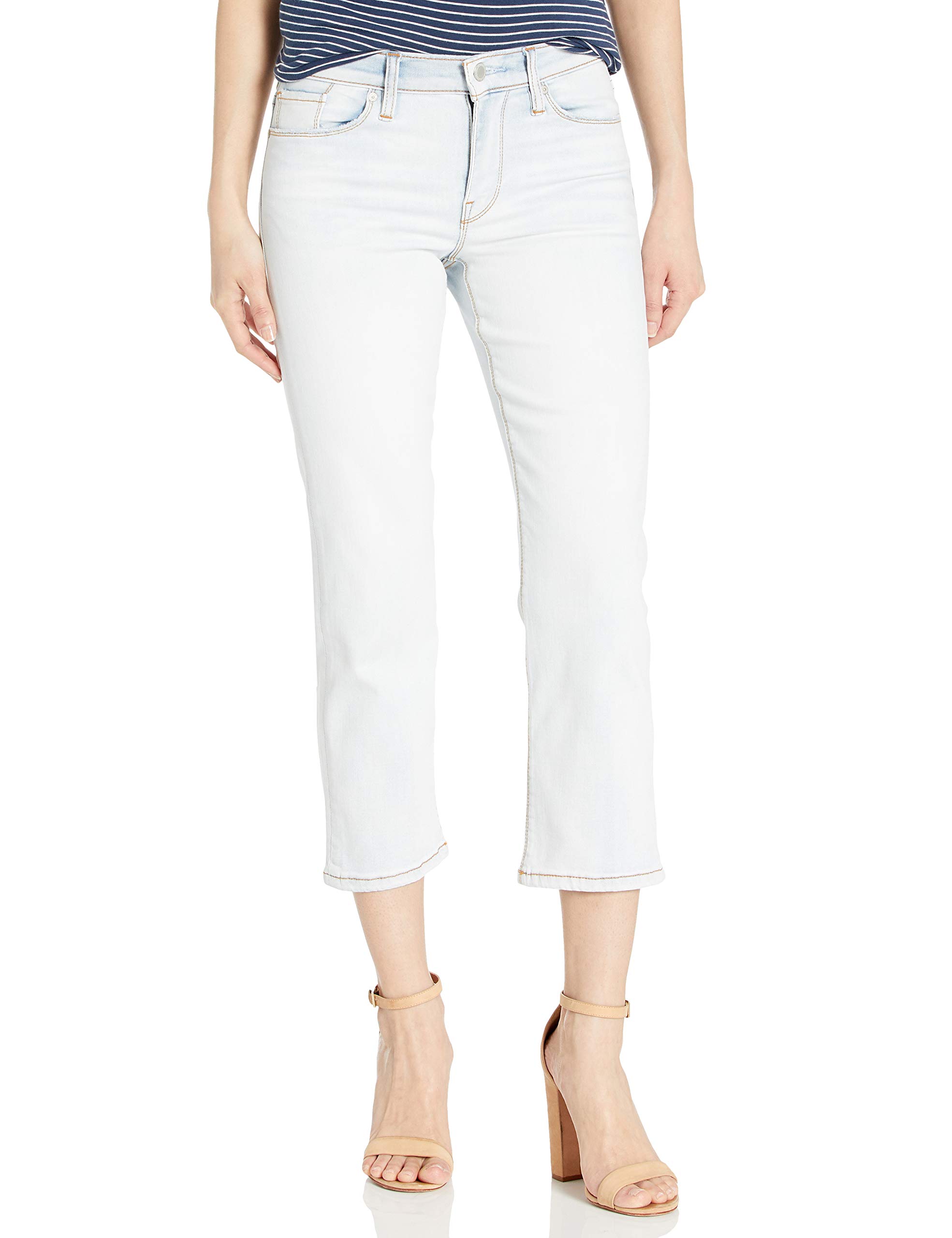}
    \caption{Sample images of majority class $casual$ archetype that are correctly predicted by IEW but failed by IFW.}
    \label{fig:examples}
\end{figure}

For the archetype experiment,  Table~\ref{table:archPeromance} shows favorable performance when using the IEW mechanism. Both weighting schemes improved the performance for some of the labels with small sample size -- $androgynous$ and $retro$. However, IFW only shows a minor improvement for less-represented labels at a cost of reducing the performance for high proportion labels -- $boho$, $casual$, and $sport$. It is important to highlight that IEW mechanism consistently improved the performance of every label. Figure~\ref{fig:examples} shows example images from $casual$ label where IFW model made incorrect predictions, but correctly classified by IEW. The model thresholds are consistently selected to ensure at least 60\% label-wise precision according to customers' needs.

Table~\ref{table:stypePeromance} shows that using IEW in sleevetype model results in a reduced performance for labels with higher number of samples. To further evaluate the impact of \textit{outlier} labels with low sample size, we compare the sleevetype model with selected labels -- bell, fitted, short, and sleeveless. Table~\ref{table:stypeSelectedPeromance} shows that removing the outlier classes improved the performance of both NW and IEW models, while using IEW resulted in greatest impact.

\begingroup
\setlength{\tabcolsep}{1.8pt} % Default value: 6pt
\renewcommand{\arraystretch}{2.8} % Default value: 1
\begin{table}[H]
    \centering
    \fontsize{6}{8}\selectfont \setlength{\tabcolsep}{0.3em}
    \caption{Comparison on the label-wise AUC performance of the three models -- NW (non-weighted), IFW (inverse-frequency as weights), and IEW (inverse-effective-size as weights) on test set for the archetype experiment. }    
    \label{table:archPeromance}
    \begin{tabular}{|| c | c | c | c| c | c| c |c |c |c |c||}
    \hline
       & \textbf{androgynous} & \textbf{boho} & \textbf{casual} & \textbf{classic}  
       &\textbf{edgy} &\textbf{glam} & \textbf{minimal} & \textbf{retro} 
       &\textbf{romantic} &\textbf{sport}\\
        \hline
          \textbf{~~~NW~~~}  &  0.061 & 0.377 & 0.682 & 0.390 & 0.183 & 0.213 & 0.222 & 0.069 & 0.421 & 0.536\\
        \hline
        \textbf{~~~IFW~~~} & 0.131 & 0.372 & 0.624 & 0.404 & 0.192 & 0.246 & 0.268 & 0.106 & 0.460 & 0.448\\
        \hline
        \textbf{~~~IEW~~~} & {\color{blue}0.188} & {\color{blue}0.572} & {\color{blue}0.763} & {\color{blue}0.604} & {\color{blue}0.443} & {\color{blue}0.526} &
        {\color{blue}0.330} & {\color{blue}0.195} & {\color{blue}0.565} & {\color{blue}0.748}\\
        \hline
    \end{tabular}
\end{table}
\endgroup

\begingroup
\setlength{\tabcolsep}{1.8pt} % Default value: 6pt
\renewcommand{\arraystretch}{2.8} % Default value: 1
\begin{table}[H]
    \centering
    \fontsize{6}{8}\selectfont \setlength{\tabcolsep}{0.3em}
    \caption{Comparing the label-wise AUC performance on test set for the sleevetype experiment. }    
    \label{table:stypePeromance}
    \begin{tabular}{|| c | c | c | c| c | c| c |c |c |c |c||}
    \hline
       & \textbf{balloon} & \textbf{batwing} & \textbf{bell} & \textbf{cap}  
       &\textbf{fitted} &\textbf{puff} & \textbf{raglan} & \textbf{short} 
       &\textbf{sleeveless} &\textbf{spaghetti}\\
        \hline
          \textbf{~~~NW~~~}  &  0.600 & 0.151 & 0.521 & 0.020 & 0.824 & 0.350 & 0.090 & 0.951 & 0.978 & 0.833\\
        \hline
        \textbf{~~~IEW~~~} & 0.526 & {\color{blue}0.161} & 0.397 & {\color{blue}0.032} & 0.792 & {\color{blue}0.613} &
        {\color{blue}0.187} & 0.929 & 0.960 & {\color{blue}0.875}\\
        \hline
    \end{tabular}
    
\end{table}
\endgroup

\begingroup
\setlength{\tabcolsep}{1.8pt} % Default value: 6pt
\renewcommand{\arraystretch}{2.8} % Default value: 1
\begin{table}[H]
    \centering
    \fontsize{6}{8}\selectfont \setlength{\tabcolsep}{0.3em}
    \caption{Comparing the label-wise AUC performance on test set for high sample size sleevetype labels. }    
    \label{table:stypeSelectedPeromance}
    \begin{tabular}{|| c | c | c | c | c | c ||}
    \hline
       & \textbf{~~~balloon~~~} & ~~~~\textbf{bell}~~~  &\textbf{~~~~fitted~~~~}  & ~~~~\textbf{short~~~~}  &\textbf{~~~sleeveless~~~} \\
        \hline
          \textbf{~~~NW~~~}  &  0.703 & 0.617 & 0.928 & 0.993 & 0.995\\
        \hline
        \textbf{~~~IEW~~~} & {\color{blue}0.756} & {\color{blue}0.665} & 0.924 & {\color{blue}0.996} & {\color{blue}0.997}\\
        \hline
    \end{tabular}
    
\end{table}
\endgroup

\subsubsection{Ablation Study}
% Further, we plot the PR-curves to compare 3 different training scenarios to test: (1) how our solution peform upon majority classes' classification, (2) how our solution performs upon slight-minority classes, and (3) how our solution performs upon severe-minority classes. 

% From Fig.~\ref{fig:majority}, we can see that by incorporating our solution, though majority classes are assigned with higher-penalties, this solution does not hurt the model performance. Instead, with a more balanced weighting distribution, the performance upon majority class is boosted. 

Further, we plot PR-curves to compare 3 different training scenarios: (1) majority classes' classification, (2) slight-minority classes' classification, and (3) severe-minority classes' classification. We aim to understand our solution's capacity in faced of different imbalanced data scenarios. 

\begin{figure}[h]
    \centering
    \includegraphics[width = 1 \textwidth]{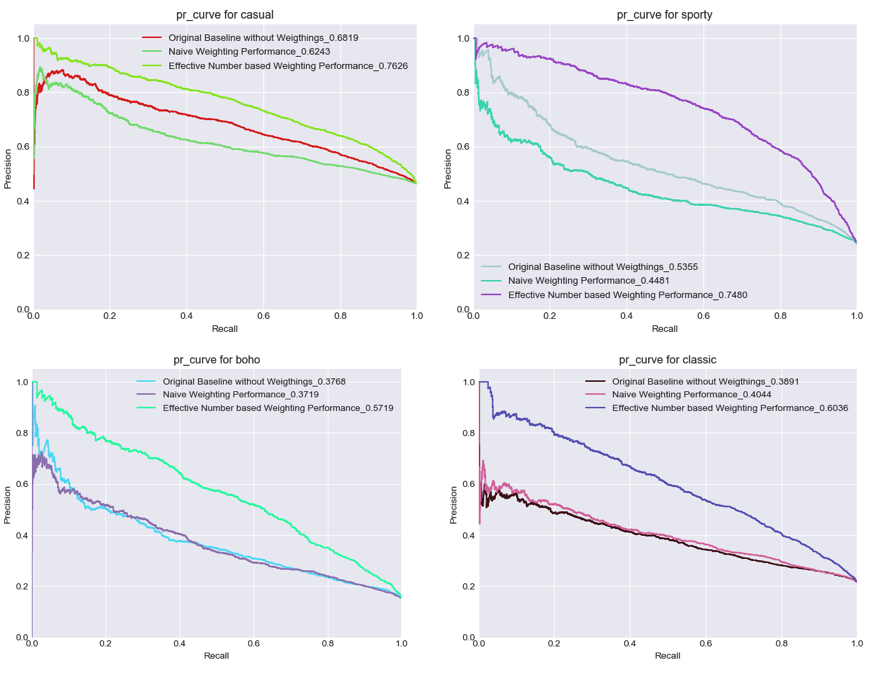}
    \caption{Model Performance on Majority Classes Comparison}
    \label{fig:majority}
\end{figure}

Fig.~\ref{fig:majority} shows that incorporating our solution does not harm the model performance for majority classes, despite assigning them higher penalties. On the contrary, the more balanced weighting distribution improves the performance of majority classes.

\begin{figure}[h]
    \centering
    \includegraphics[width = 1 \textwidth]{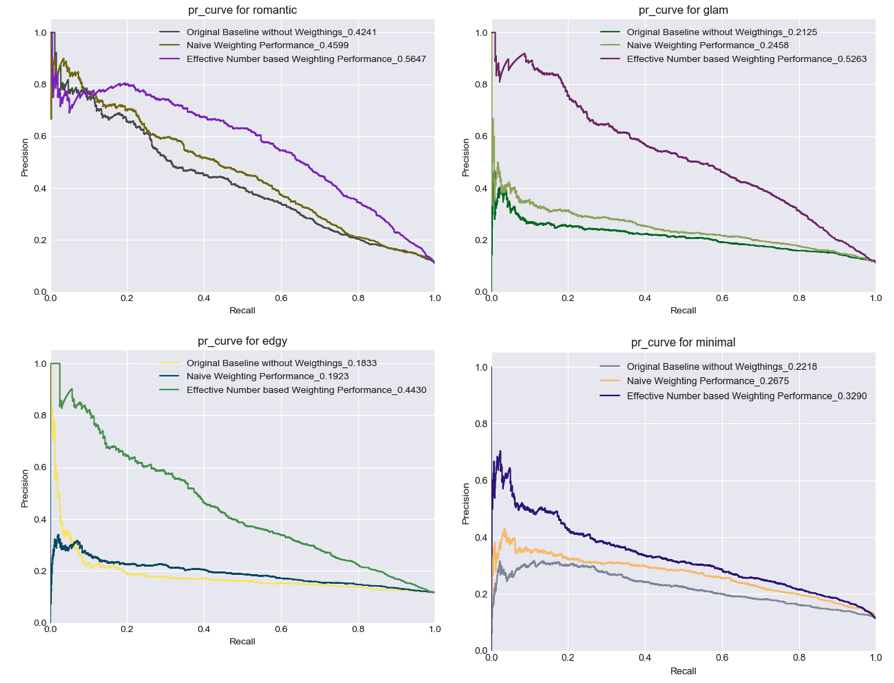}
    \caption{Model Performance on Reasonable Minority Classes Comparison}
    \label{fig:minority}
\end{figure}

\begin{figure}[h]
    \centering
    \includegraphics[width = 1 \textwidth]{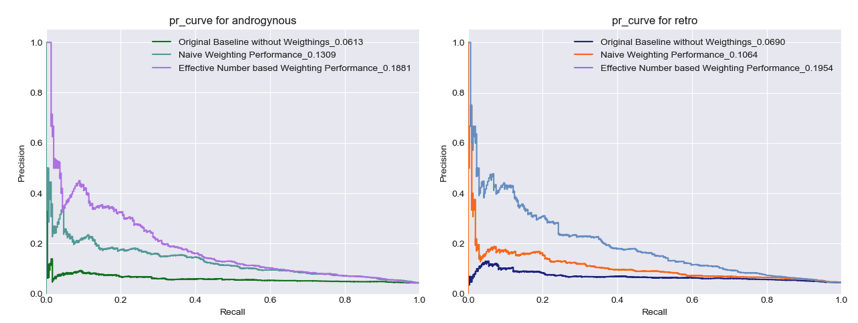}
    \caption{Model Performance on Reasonable Minority Classes Comparison}
    \label{fig:badminority}
\end{figure}

More importantly, our solution is extremely effective towards the minor-minority classes. From Fig.~\ref{fig:minority} it is clear to see the improvement is significant. On the other hand, from Fig.~\ref{fig:badminority}, we see that when the samples of a certain class is inherently too under-numbered, with this numerical remedy, our solution still fails to yield a significant improvement. This still indicates that data itself plays an extremely important role despite all other kinds of training techniques, model architectures, and numerical adjustments. 

%% file: sec5_conclusion.tex
\section{Conclusion}
Overall, using effective number of samples as the weighting scheme enables us to obtain a well-balanced weight for each class, which could better represent the true population's label distribution through mathematical expectation. Experimental results on fashion attribute classification show significant boost on the multi-label classification model performance using introduced weighting scheme on both the majority classes and the minority classes. Moreoever, this discovery exhibits promising potential for widespread application across various industrial domains that rely on pattern recognition. Particularly noteworthy is its suitability for addressing challenging scenarios characterized by the presence of edge cases, making it an invaluable asset in such contexts~\cite{mei2017mechanics,mei2018erratum}. With further investigation, we identify some limitation of this method which requires a lower bound of sample size of the minority classes. In future work, we will explore possible mitigation like some cut-off line of the least amount sample size for each label such that the effective number of samples based weighting mechanism could operate consistently with improved performance. Further, the idea of effective sample size is solely based on sample size and decoupled with the modality of input data. A combined model could help assess the complex task of evaluating the limits of machine learning datasets.